# On mistakes we made in prior Computational Psychiatry Data driven approach projects and how they jeopardize translation of those findings in clinical practice

Milena Čukić Radenković [1,2,*], Dragoljub Pokrajac[3], Viktoria Lopez[2]

[1]Department for General Physiology and Biophysics, University of Belgrade, Belgrade, Serbia
[2]Instituto de Tecnología del Conocimiento, Universidad Complutense de Madrid, Spain
[3]Business and Finance Division, Delaware State University, USA

**Abstract**

In this work we aimed at comparing our findings in depression detection task with methodologies applied in present literature. Previously we showed that when electrophysiological signal (in this case electroencephalogram, EEG) is characterized by nonlinear measures, any of seven most popular classifiers yields high accuracy on the task. Following every step we done in this process we compare it with other researchers' practice and comment on other findings mainly from analysis of electrical signals or nonlinear analysis showing what would be optimal for further research. We focused on discussing various mistakes and differences that could potentially lead to unwarranted optimism and other misinterpretation of results. In Conclusion we summarize recommendation for future research in order to be applicable in clinical practice.

## Introduction

Current clinical psychiatry is lacking objective biochemical or electrophysiological tests used for diagnosis unlike other medical disciplines. To diagnose depression, clinician will typically rely on the self-report from the patient and his experience in applying DSM manual, which is standardized list of symptoms to be checked in every case (in order to be qualified as a certain disorder). It is perfectly possible that two persons diagnosed with the same disorder have not overlapping symptoms, and that one person can have two distinct diagnosis. If someone has more than three episodes of depression, that is considered to be recurrent depression (after every episode the probability of the next one is doubling). This is particularly heard to treat and manage therapy which is ongoing through person's whole life. Apart from obsolete diagnostic, all antidepressants have serious side-effects, the waiting lists are very long (in Nederland they are between 6 and 9 months long) and the therapy can last for years or even decades. It is reported than only 11 - 30% of patients are improving in the first year of therapy (Rush et al., 2008). It was also found that when people report symptoms online (Mechanical Turk) it can be showed that the trained professional examiner is no better at applying DSM IV (Berinsky et al., 2015). Computational psychiatry, and in particular its data-driven approach (typically using machine learning) is aiming at tackling this problem, since depression is becoming one of the most frequent causes of disability worldwide (WHO, WEF) with an onset as early as 12 years.

Physiological complexity, like machine learning is not so novel concept, but those who are specialized in medicine, and in particular in psychiatry it is considered to be novel. The reason why any electrophysiological signal (like EEG, ECG or EMG) is still analyzed by reductionistic approach, like Fourier's analysis, is probably because that practice is standard in electrophysiology for so long, and nonlinear analysis is not (it is mainly reserved for physiological research) (Klonowski et al., 2007). With 'nonlinear analysis' we mean nonlinear measures originating from Complex Systems Dynamics Theory (or more popular, Chaos theory). Electrophysiological signals are nonlinear, nonstationary and noisy (Klonowski, 2006, Goldberger et al., 2001). Electroencephalogram, which is generated by brain cells (and brain is one of the most complex dynamical systems we know) and do represent very complex dynamics can hardly be understood by application of classical (linear) methods of analysis. Still the Fourier is the only analysis applied (although if the signal is decomposed by wavelet or cosine transform the distortion of original signal is similar) and standard sub-bands are still in use although there is no evidence that those have any physiological significance nor meaning (Bacar et al., 2015).

In many published research it was confirmed that an elevated physiological complexity is present in depression (for review see De la Torre-Luque and Bornas, 2017). Our understanding of that, based on previous research and present literature is that due to decreased connectivity within deeper structures important for depression, an elevated excitability on cortex can be detected (De Kwaasteniet et al., 2013; Kim et al., 2013). That difference can be used for detection, and EEG is way more cheaper way of recording (it is basically the oldest neuroimaging method) present in modern hospitals.

In the last eight years, the number of research utilizing some form of machine learning on EEG dataset to detect the depression or predict the outcome of treatment, related to depression is flourishing. We are interested here to comment only the work where nonlinear features were used for further classification task (patients diagnosed with depression and healthy controls), and to be able to compare it with our results we opted to mention only those which used resting-state EEG as signal for analysis (although there are many other using for example fMRI, MRI or other methods based on which they do the classification or prediction).

In this work we are presenting various methodological combinations to accurately solve depression classification task (differentiation of people who have depression and healthy controls), including our own results. We are going to compare and discuss already detected mistakes and possible misinterpretations, as well as other possible problems, and offer some solutions to improve that practice in the future. At the end we are proposing some practices which we believe will lead to more realistic results, therefore avoiding unwarranted optimism in machine learning applications in psychiatry.

**Method**

The aim of our research was to show that working with minimally preprocessed resting state EEG, without using standard sub-bands, but applying broad-band analysis and nonlinear characterization of signal, all popular classifiers can yield decent accuracy in the task.

Based on research with other electrophysiological signals we focused on two nonlinear measures which are computationally fast, namely Higuchi Fractal Dimension (HFD) and Sample entropy (SampEn). Higuchi published his method in 1988 (Higuchi, 1988) and Richman and Moorman published algorithm for improvement in Aproximate etropy (ApEn), SampEn in 2000 (Richmann and Moorman, 2000). We used both algorithms to write in-house made algorithms in Java programming language. Both measures were calculated directly on time series.

To determine the linear dependence of the EEG features, we utilized correlation analysis. We applied PCA to determine the influence of linear feature extraction on classification accuracy. We examined various classification algorithms, from the simplest and linear in parameters to highly non-linear: Logistic regression (LR), Support vector machines (SVM) both with the linear and polynomial kernel, Multilayer perceptron (MP), Decision tree (DT), Random forests (RF) and Naïve Bayes (NB).

After calculation of nonlinear features in Java, the classification was performed using Weka software (Weka v. 3.8, University of Waikato). Principal component analysis was computed using Matlab (Matlab v. R2015b, Mathworks). Statistical analysis was performed using SPSS software (IBM SPSS Statistics 20).

*Data acquisition*

The data used for our research were recorded at Institute for Mental Health in Belgrade, Serbia. The subjects were 16 patients (12 women and 4 men) suffering from recurrent DD, 24 to 68 years old (mean 31.53, SD 10.21). As a control we used EEG records of 20 age – matched (mean 30.14, SD 8.94) healthy controls (10 males, 10 females) with no previous history of any neurological or psychiatric disorders, recorded at Institute for Experimental Phonetic and Speech Pathology in Belgrade, Serbia.

The participants from both groups were all right-handed, according to Edinburgh Handedness Inventory. All the participants were informed about the protocol of the experiment and signed informed consent; the protocol was approved by Local Ethical Committees of the participating institutions. All procedures performed in studies involving human participants were in accordance with the ethical standards of the institutional and/or national research committee and with the 1964 Helsinki declaration and its later amendments or comparable ethical standards. All the participants were on the pharmacological therapy prescribed by their psychiatrist. All the diagnoses were made according to ICD-10 scale. The patients' EEGs were recorded after the visit to a recommended specialist, a psychiatrist. EEG was recorded in the resting state with 10-20 system using NicoletOne Digital EEG Amplifier (VIAYSYS Healthcare Inc. NeuroCare Group), with closed eyes without any stimulus. EEGs were obtained from 19 electrodes in a monopolar montage (Electro-cap International Inc. Eaton, OH USA) – Frontal regions: Fp1, Fp2, F3, F4, F7, F8 and Fz; Central regions: C3, C4, and Cz; Parietal regions: P3, P4, and Pz; Temporal regions: T3, T4, T5, and T6; Occipital regions: O1 and O2. Odd numbers are related to the left and even numbers to the right hemisphere. A sampling rate was 1 kHz; the resistance of electrodes was less than 5 kΩ, bandpass was 0.5-70 Hz. For the control group, the same setup was used, but on the Nihon Kohden Inc. apparatus.

Every recording lasted 3 minutes. Subjects were instructed to reduce any movement, staying in a comfortable sitting position with eyes closed. After recording, we had to discard records of two subjects from further analysis due to low voltage EEG in one male participant's case and epileptic seizure very close in time from the recording in one female participant. So, we used records from 14 patients with depression and 20 healthy controls for this study. Artifacts were carefully inspected by an expert and epochs for analysis were taken from artifact-free traces.

*Fractal analysis*

Fractal dimension (FD) of a time series is a measure of its complexity and self-similarity in the time domain. FD is a number in the interval [1, 2]. Generally, higher self-similarity and complexity result in higher FD [Eke et al, 2002.]. The fractal dimension of EEG was calculated by using Higuchi's algorithm [36] demonstrated to be the most appropriate for electrophysiological data [Esteller 1999, Castigloni 1998]. This method works directly in time domain, gives a reasonable estimate of the fractal dimension even in the case of short signal segments and is computationally fast (since it does not attempt to reconstruct the strange attractor). We performed the Highuchi's algorithms with the maximal scale (Higuchi, 1988) $k_{max}=$ 8 shown to perform the best for this type of signals [37]. Fractal dimensions were calculated for each electrode for the duration of signal (the epoch of recorded EEG), and the calculated values formed ensembles for further analysis.

*Sample entropy analysis*

Sample Entropy (SampEn) was computed according to the procedure proposed by [38]. SampEn estimates the signal complexity by computing the conditional probability that two sequences of the given length, *m*, similar for *m* points, remain similar within tolerance *r* at the next data point (when self-matches are not included). Mathematically, SampEn is the negative natural logarithm of the conditional probability that two sequences similar for the *m* points remain similar at the next point. Since SampEn measures the irregularity of the data (the higher values, the less regular signal) that is related to signal complexity [Glodberger et al 2001.]. According to the previous study [39], we used a tolerance level of *r* = 0.15 times the standard deviation of the time series and *m* = 2.

Both HFD and SampEn resulted in 19 features (the number of electrodes). We merged them to get 38 features for further machine learning analysis. To examine statistical difference between the groups we used MANOVA (SPSS Statistics version 20.0, SPSS Inc, USA), followed by posthoc Bonferroni tests, where appropriate.

## CLASSIFIERS

In this study, we compare the performance of several classifiers to discriminate between patients with DD and controls (a two-class problem). A classifier can be formally defined as mapping between a feature vector $\mathbf{x} = [x_1 \dots x_k]$ and a class label $c \in \{Patient, Control\}$. We utilize classifiers implemented in Weka software [34] with their default parameter values. All classifiers are applied on normalized features. The normalization is performed by subtracting sample means and dividing by sample standard deviation such that the inputs of algorithms have zero means and unit standard deviation.

*Naïve Bayes*

Naïve Bayes classifier [40] is a variant of maximum a posteriori classifier that outputs the class with the highest probability given the observed feature values. This posterior probability is calculated using the Bayes' theorem [41]. The observed attributes are assumed independent. Hence, the posterior probability conditioned by the set of attributes is calculated as the product of posterior probabilities conditioned by each feature:

$$P(x_1, \dots, x_k | c_i) \sim P(c_i) * \prod_{j=1}^{k} P(c_i | x_j) \qquad (1)$$

Here, $c_i$ is a class label of *i*-th example, $k$ it the number of features and P denotes probability. In this study, these posterior probabilities are assumed to have normal distributions [42].

*Logistic regression*

Logistic regression [43] estimates the class conditional probability using a linear combination of features and logistic regression function $f(y)$:

$$f(y) = \frac{1}{1+e^{-y}}. \qquad (2)$$

Specifically, in the considered two-class problem:

$$P(c_i = Patient | x_1, \dots, x_k) = \frac{1}{1+e^{-\left(\beta_0 + \sum_{i=1}^{k} \beta_i x_i\right)}}. \qquad (3)$$

The coefficients $\beta_0, \beta_1, \dots, \beta_k$ are estimated using the LogitBoost algorithm [44].

*Multi-layer perceptron (MLP)*

Multi-layer perceptron is a non-linear classifier that estimates a class based on the value of its output processing unit (so-called output neuron) [44], [45]. For two-class problems (such as patients vs. controls considered in this study), multi-layer perceptron has only one output and classification is performed by its thresholding. The output neuron applies a non-linear transfer function to a linear combination of the outputs of so-called hidden neurons. In the case of MLPs with one hidden layer, used in this study, each hidden neuron computes a linear combination of the MLP input features and applies the transfer function of it. We utilize a sigmoid transfer function defined by Eq. 2. Hence, the utilized MLP can be considered as a generalization of logistic regression.

The MLP is uniquely determined by the number of hidden neurons and the coefficients utilized to calculate linear combinations of neuron inputs. We utilize MLPs with $\left\lceil\frac{k+1}{2}\right\rceil$ hidden neurons and back propagation algorithm with momentum [45] to determine the coefficients. The coefficients are learned through 500 iterations (epochs) and we use learning rate and momentum values of 0.3 and 0.2 correspondingly (default values from Weka software).

*Support vector machines (SVM)*

SVM for a two-class problem performs classification are based on partitioning a feature space into two disjoint subspaces, each corresponding to one class. The subspaces are separated by a decision boundary. The decision boundary is uniquely determined by a subset of data – support vectors. SVMs produce maximal margin classifier that maximizes the distance between the decision boundary and the data closest to it (the support vectors). SVMs produce linear decision boundaries in a transformed space determined by a kernel function. In this study, the linear decision boundaries in the original feature space are obtained using the linear kernel defined as dot product of feature vectors:

$$K(\mathbf{x}, \mathbf{y}) = \mathbf{x}^T \mathbf{y} \qquad (4)$$

We also utilize SVMs with polynomial kernel:

$$K(\mathbf{x}, \mathbf{y}) = (\mathbf{x}^T \mathbf{y})^p, p = 2 \qquad (5)$$

that results in non-linear decision boundaries at the original feature space.

To deal with a potential case in which classes are not separable, SVMs introduce a regularization constant $C$ that penalizes samples that cannot be separated. Formally, learning SVM for a two-class problem can be represented as the following optimization problem [46]:

$$\min_{\boldsymbol{\alpha}} \left( \sum_{i=1}^{N} \alpha_i - \sum_{i=1}^{N} \sum_{j=1}^{N} \alpha_i \alpha_j c_i c_j K(\mathbf{x}_i, \mathbf{x}_j) \right) \qquad (6)$$

Where

$$\sum_{i=1}^{N} \alpha_i c_i = 0; \alpha_i \geq 0, \alpha_i < C, i = 1, \dots, N \qquad (7)$$

Here, $N$ denotes the size of the training set, and $c_i$ is a binary class label (1 if a sample is patient, -1 otherwise).

A new data sample, $\mathbf{x}_{new}$, is classified according to:

$$sign\left(\sum_{i::\alpha_i>0} \alpha_i c_i K(\mathbf{x}_i, \mathbf{x}_{new}) + \frac{1}{N_s}\left(c_i - \sum_{j::\alpha_j>0} \alpha_j c_j K(\mathbf{x}_j, \mathbf{x}_i)\right)\right) \quad (8)$$

where $N_s$ is the number of support vectors.

To determine parameters of SVM, we utilize a sequential minimal optimization algorithm [47]. The regularization coefficient is set to 1.

*Decision trees*

Decision trees [48], alike SVMs, feature partition space in regions corresponding to classes. However, the regions are not necessarily connected, and the decision boundaries are piecewise linear functions. The technique associates the whole feature space with a root node of a tree and, in the case of the C4.5 algorithm utilize in this study, recursively partitions space by choosing a feature that provides the highest information gain. The partition stops when the minimal number $N_{min}$ of samples per node of a decision tree is reached (we utilize $N_{min}=2$). In the pruning phase, based on the estimation of the classification error (using a confidence level here set to 0.25) the complexity of the model may be reduced and its generalization capacity thus improved [49].

*Random forests*

Random forests classifier utilizes an ensemble of unpruned trees [50]. The classification is performed by combining classes predicted by ensemble members. Unlike C4.5 algorithm, in each node of a tree, a random subset of the features is considered for partitioning. In this study we utilize ensembles with 100 members and consider int($\log_2 k$)+1 random features for each split.

*Evaluation of classifier's performance*

The classification accuracy is evaluated through a cross-validation procedure; the dataset is split into K subsets of approximately equal size; K-1 subsets are used to determine a classification model and the remaining subset to evaluate the classifier. This procedure is repeated K times such that a classifier is evaluated in each subset. In this study, we used K=10.The classification accuracy is assessed through overall accuracy—the percentage of correctly classified samples—and using the area under the ROC curve (AUC). The overall accuracy of useful classifiers in two-class problems ranges from 50% to 100%. The ROC curve is created by plotting true positive rate (the proportion of samples with DD that are detected as such) vs. false positive rate (the ratio of the total number of controls incorrectly detected as with DD and the total number of controls). AUC ranges from 0.500 (for a classifier that randomly guesses a class) to 1.000 (for an ideal classifier).

*Feature extraction*

To reduce the dimensionality of the feature set and decorrelate the features, we utilize PCA (Joliffe, 2002) in order to obtain principal components (PCs). PCA normalizes original features by subtracting sample means and calculates sample covariance matrix of such data. The eigenvalue decomposition of the sample covariance matrix yields eigenvectors that are used to linearly project the original feature vectors **x** into the uncorrelated transformed feature vectors $\mathbf{z}=[z_1,\ldots,z_k]$. We utilize m<k features $z_1,\ldots,z_m$ corresponding to *m* largest eigenvalues of the sample covariance matrix. We define percentage of the explained variance by first *m* PCs as

$$\text{Explained Variance } = \frac{\sum_{i=1}^{m} \hat{\sigma}_{z_i}^2}{\sum_{i=1}^{k} \hat{\sigma}_{x_i}^2} * 100\% \qquad (9)$$

where $\hat{\sigma}_\cdot^2$ denotes a sample variance of a corresponding variable.

## Results

Both HFD and SampEn showed significant difference between patients with depression and healthy controls. Since this paper is focused on classification results, we are not reporting those here (it stays in domain of physiological complexity). After decorrelation between HD and SampEn we used them as features for further classification and we also used PCA as method of decreasing the dimensionality of the problem.

**Table 1**. shows classification results of different classifiers that use various numbers of principal components. The principal components were computed on a dataset containing HFD and SampEn features and were normalized to have zero mean and unit standard deviation. The variance of the features explained by the corresponding principal components is also shown.

CLASSIFICATION RESULTS FOR DIFFERENT CLASSIFIERS AND DIFFERENT NUMBER OF PRINCIPAL COMPONENTS OF THE FEATURES FROM SAMPEN AND HFD SETS (ČUKIĆ ET AL., 2020)

| Number of principal components | 1 | | 2 | | 3 | | 10 | |
|---|---|---|---|---|---|---|---|---|
| Explained variance | 87.53% | | 94.25% | | 95.54% | | 98.86% | |
| Classifier | Accuracy | AUC[b] | Accuracy | AUC | Accuracy | AUC | Accuracy | AUC |
| Multilayer perceptron | 92.68% | 0.983 | 92.68% | 0.943 | 92.68% | 0.950 | 95.12% | 0.994 |
| Logistic regression | 90.24% | 0.981 | 90.24% | 0.950 | 90.24% | 0.945 | 95.12% | 0.929 |
| SVM[a] with linear kernel | 85.37% | 0.857 | 82.92% | 0.833 | 85.37% | 0.857 | 90.24% | 0.905 |
| SVM with polynomial kernel (p=2) | 73.17% | 0.738 | 68.29% | 0.690 | 85.37% | 0.857 | 95.12% | 0.952 |
| Decision tree | 92.68% | 0.894 | 92.68% | 0.894 | 90.24% | 0.933 | 90.24% | 0.933 |
| Random forest | 87.80% | 0.981 | 87.80% | 0.981 | 95.12% | 0.987 | 95.12% | 0.996 |
| Naïve Bayes | 95.12% | 0.983 | 97.56% | 0.981 | 95.12% | 0.988 | 95.12% | 0.988 |
| Average accuracy | 88.15% | | 87.45% | | 90.59% | | 93.73% | |

[a] SVM - Support Vector Machines, [b] AUC - area under the curve (related to Receiver Operating Characteristic – ROC curves)

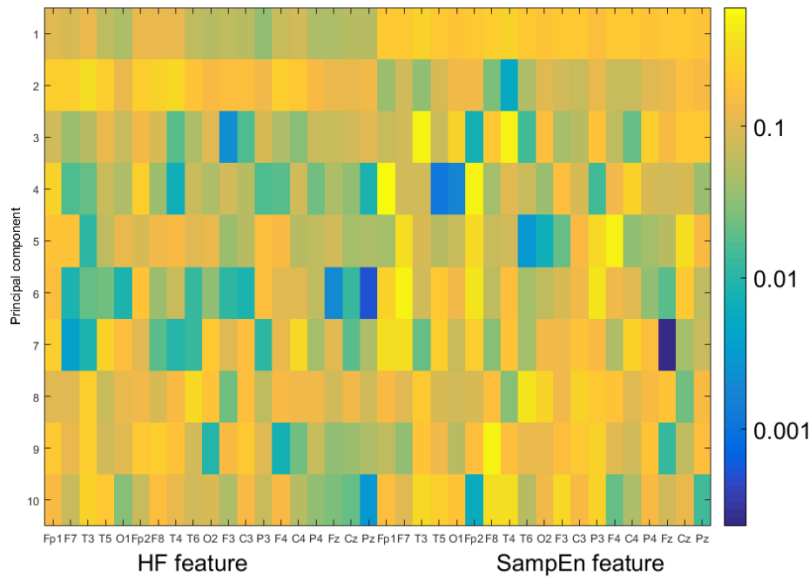

**Figure 1**. Absolute values of principal components loads for first 10 principal components. Each row contains indicates the coefficients multiplying corresponding non-linear feature in order to generate a principal component. From Čukić et al., 2020.

**Discussion**

The major finding of our experiment is that the extraction of non-linear features linked to the complexity of EEG signals can lead to high and potentially useful separation between signals recorded from control subjects and patients diagnosed with depression. In addition, we demonstrated that Higuchi's fractal dimension (HFD) and sample entropy (SampEn) could be used as suitable features for different machine learning classification techniques. Between the two SampEn showed to yield better performance.

In line with previous findings (Bachmann et al., 2013; Ahmadlou et al., 2012) about measures used for characterization of EEG, HFD detected increased complexity of EEG recorded from patients with DD in comparison to healthy controls. We showed that SampEn also effectively discriminates these two EEG signals. Such significantly different physiological complexity indicated by both measures can alone serve to distinguish patients' with depression EEG from healthy subject's EEG. Our aim with the application of different methods for distinguishing depressed from healthy subjects was not to improve the accuracy of classifiers but to show that with sufficiently good features (i.e., nonlinear measures) a good separation is possible. Further, by using PCA, we strived to demonstrate that such measures may be dependent on each other. Hence the good classification results are possible with a small number of extracted principal components. Therefore we demonstrated separability of the data with all methods utilized here. Note that we examined classification methods with a range of underlying paradigms and complexity; the methods belong to statistics and machine learning. Even the simplest methods, such as logistic regression (widely accepted in the medical community although not a classification method in the strict sense) provided excellent classification accuracy. This is in agreement with previous findings (Pokrajac et al., 2014) about the importance of proper feature extraction.

Our results show that using only the first principal component, it is possible to achieve a classification accuracy of up to 95.12% (Table 1). The best performance was achieved using the Naïve Bayes method (in this case, when only one feature is utilized, the assumption of feature independence is automatically satisfied). The classification accuracy generally increases with the number of principal components used at classifiers' inputs (the average accuracy of all classifiers is 88.15% with 1 and 93.73% with 10 principal components used). Since the data are close to linearly separable, SVM with linear kernel resulted in relatively high accuracy (85.37%) when only one PC is used. The accuracy further increased to 90.24% with 10 PCs. This can be in part explained by Cover's theorem [56] that data in higher dimensional spaces tend to be more likely linearly separable. It is interesting to observe that with a small number of PCs, linear SVM outperformed SVMs with the polynomial kernel; our hypothesis is that in such cases the inherent non-linearity of polynomial SVMs cannot be fully exploited. Also, this is an indication that the choice of the kernel for SVM may play a significant role in the classification performance of the model. The random forest method benefited from a larger number of utilized principal components, since the method is based on randomly choosing one from a set of available features to split a decision tree node (when the number of PCs used is small, the set of available features is small).

It is important to remeber that the goal of our study was not to evaluate the optimal classification accuracy of the discussed classifiers, but to demonstrate the discriminative power of non-linear features. Hence, we utilized default parameters of the classification algorithms from Weka software and did not try to optimize those (Unnickrishan et al., 2016). The generalization properties of the classifiers are measured using a standard 10-fold cross-validation technique. Since this was only a pilot study, and the repetition of the results is needed on larger data set prior to making a final conclusion about class separability and the potential application of the classification techniques for diagnostic purposes.

When we compare our methodology with present body of evidence, there are several important stages of analysis that can be learned from. First of all, majority of previously published work was done on small to modest samples, our study included (although we stated it was pilot study).

Ahmadlou performed classification task with the idea to compare two algorithms for calculating the fractal dimension, Higushi's and Katz's (Ahmadlou et al. 2012). Esteller et al., (2001) showed that KFD is more robust to noise compared to Petrosian's and Higuchi's algorithm, but others showed (Raghavendra and Narayana 2009; Castiglioni, 2010) that KFD is dependent on the sampling frequency, amplitude and waveforms, which is a disadvantage in analyzing bio signals. The disadvantage of HFD is, according to Esteller (2001) that it is more sensitive to noise than KFD. The sample for Ahmadlou study comprised of 12 non-medicated MDD patients and 12 healthy controls. Following studies also had bigger but still modest samples like in

(Puthankattil et al., 2012; Faust et al., 2014; Acharya et al., 2015; Bairy et al. 2015). Puthankattil had 30 depression and 30 controls, 60 together; Faust and Bairy also, and Ahcarya opted to use just 15 +15. Hosseinifard (Hosseinifard et al., 2014) analyzed sample of 90 persons (45D+45HC), and Bachmann analyzed 33D+30HC (Bachmann et al., 2018). Liao and colleagues had the same sample size as Ahmadlou (Liao et al., 2017). Based on previous knowledge from data mining, we know that only on bigger samples we can truly test our models, small samples are leading to so-called 'unwarranted optimism' or 'inflated optimism' (too high to be true accuracy), apart from the very present practice in published papers that the proper validation was not performed.

Among studies we mentioned, only Hosseinifard succeeded in having non-medicated patients diagnosed with depression, and Bachmann group have chosen to analyze the sample consisting of female participants only. When we are talking about sampling frequency contrary to Hosseinifard and our own research (1kHz), all other researchers used sampling rate of raw EEG to be 256Hz. Also, Hosseinifard was the only one to analyze all the traces from 19 electrodes (as well as our group) while others opted to use just 1 (Bachmann), 4 (Puthankattil, Faust, Acharya, Bairy), or 7 (Ahmadlou) claiming that it would be enough for detection and that clinicians would most probably prefer single electrode detection (Bachmann). From our results it is clear that all the electrodes are giving their contribution to overall result, as shown on Figure 1. Therefore, we strongly support that researchers analyze all the traces they have. They can deal with dimensionality later, but at this stage we think it is important to include all the electrodes.

The issue with sub-bands is that although they are in use for very long time, no one still did not confirm their physiological meaning nor significance (Basar et al., 2001); why not using broad-band? From above mentioned studies performing depression classification task that was part of the analysis only in our group work and in Hosseinifard's. We based our decision on previous work of Goldberger, Peng, Lipsiz, Pincus and others who repeatedly showed that the non-filtered signal is most information rich, when we are performing any kind of nonlinear analysis (Goldberger et al., 2001; Pincus, 1998). The complex intrinsic dynamics of an electrophysiological signal can be destroyed with too much preprocessing, and that is why we think this is important question.

It is also interesting how researchers are handling feature extraction and feature selection; in our case HFD and SampEn was the former and PCA the later. While feature extraction equals creating the features, feature selections' task is to remove features that are irrelevant or redundant. In Ahmadlou group case the features were fractal dimensions calculated by two distinct algorithms and ANOVA was used to extract those which were relevant, meaning able to differentiate within groups (Ahmadlou et al., 2012). Puthankattil and team extracted 12 features from prior eight level multiresolution decomposition method of discrete wavelet transform to create feature, i.e. wavelet entropy (Puthancattil et al., 2012), while Faust used five different entropies as feature, and Student's t-test to evaluate features them (Faust et al., 2014).

Hosseinifard and colleagues used spectral power together with HFD, Correlation dimension and Lapynov exponent (LLE) as features for EEG (Hosseinifard et al., 2014). Acharya applied 15 different measures both spectral and nonlinear measures for feature extraction: fractal dimension

(Higuchi fractal dimension, HFD), Largest Lyapunov exponent (LLE), sample entropy (SampEn), DFA, Hurst's exponent (H), higher order spectra features (weighted Centre of bispectrum, W_Bx, W_By), bispectrum phase entropy (EntPh), normalized bispectral entropy (Ent1) and normalized bispectra squared entropies (Ent2, Ent3), and recurrence quantification analysis parameters (determinism (DET), entropy (ENTR), laminarity (LAM) and recurrent times (T2)). They were ranked using the t value. After that authors formulated Depression Diagnosis Index taking into account only LAM, W_By and SampEn (Acharya et al., 2015). Hosseinifard used a genetic algorithm (GA) for feature selection. Bachmann used spectral measure SASI, but also calculated Higuchi fractal dimension (HFD), detrended fluctuation analysis (DFA) and Lempel-Ziv complexity as features (Bachmann et al., 2018).

In their paper, Liao et al. (2017) proposed a method based on scalp EEG and robust spectral spatial EEG feature extraction based on kernel eigen-filter-bank common spatial pattern (KEFB-CSP). They first filter the multi-channel EEG signals (30 electrodes traces) of each sub-band from the original sensor space to new space where the new signals (i.e., CSPs) are optimal for the classification between MDD and healthy controls, and finally applies the kernel principal component analysis (kernel PCA) to transform the vector containing the CSPs from all frequency sub-bands to a lower-dimensional feature vector called KEFB-CSP.

When we try to compare all mentioned studies, it is clear that the task is challenging, from methodological point of view. For example, it is enough that we calculated HFD by utilization of the same algorithm but opted to use different kmax from someone else, and, final results are really hard to compare. Also, while experimenting with algorithm written in various programming languages (Matlab, C++, Phyton, Java, Go) we noticed that the time of execution varies greatly and opted to use Java with the fastest execution below 10ms (for one particular time series extracted from the raw EEG). We need to mention here that we used resources from Physionet.org to compare their algorithms with ours in order to acquire high accuracy and be able to compare the results on their samples.

Some of the studies we mentioned have very low reproducibility; for example, Bairy did not even mention what algorithm they used for calculating the fractal dimension (Bairy et al., 2015). While Ahmadlou used Enhanced Probabilistic Neural Networks (EPNN), Puthankattil used artificial feedforward neural network, and Hosseinifard used K-nearest neighbors (KNN), Linear Discriminant Analysis (LDA) and Linear Regression (LR) classifiers. Faust used Gaussian Mixture Model (GMM), Decision Trees (DT), K nearest neighbors (KNN), Naïve Bayes Classifier (NBC), Probabilistic Neural Networks (PNN), Fuzzy Sugeno Classifier (FSC) and Support Vector Machines (SVM), while Acharya used SVM only. It is interesting that many researchers did not even reported the method of validation, like Acharya (Acharya et al., 2015). In Bachmann (2018) used logistic regression with leave-one-out cross-validation. Ten-fold cross validation was used also in our work.

Based on HFD (which outperformed KFD) Ahmadlou group obtained high accuracy of 91.3%. Puthankattil obtained performance of artificial neural networks so it resulted in an accuracy of 98.11% (normal and depression signals). Hosseinifard reported classification accuracy the best in the alpha band for LDA and LR both reaching 73.3% (the worst was KNN in delta and beta and

LDA in the delta with 66.6%). The best accuracy in the experiment was obtained by LR and LDA classifiers. Conclusion was that 'nonlinear features give much better results in the classification of depressed patients and normal subjects' contrary to the classical one. Also, they concluded that depression patients and controls differ in the alpha band more than other bands, especially in the left hemisphere (Hosseinifard et al., 2014). The accuracy was 99.5%. Acharya was obtaining the accuracy higher than 98% . Faust (2014), applied ten-fold stratified cross-validation; the accuracy was 99.5%, sensitivity 99.2%, and specificity 99.7%. Contrary to Hosseinifard they claim that the EEG signals from the right part of the brain discriminate better the depressive persons (Faust et al., 2014). Bairy reported an accuracy of 93.8% (Bairy et al., 2015). We cannot say whether internal or external validation was performed. Liao (2017) achieved 80% accuracy with their KEFB-CSP While Bachmann (2018) obtained maximal accuracy of 85% with HFD and DFA, but also HFD and LZC, and for only one nonlinear measure maximal 77%. Average accuracy among classifiers obtained in our work ranged from 90.24% to 97.56%. Among the two measures, SampEn had better performance.

Our comparison show that although above mentioned publications used various combination of features and machine learning models, they overall have reached high accuracy in classifying depressive and healthy participants based on their resting-state EEG. Although their direct comparison is challenging, the common denominator for all presented studies can be summarized as comparing methodological steps inevitable in this kind of research where certain features previously shown to be characteristic for depression were used to feed classifiers.

Based on present body of literature the characteristic of depression is an elevated complexity of EEG when compared to healthy peers (for review see, de la Torre-Luque, 2017). In present literature using fMRI, fractional anisotropy (FA) (de Kwaasterniet et al., 2013; Vederine et al., 2011) and graph theory applications on EEG signal (Kim et al., 2013) changes in functional connectivity characteristic in depression are demonstrated. That is consequently reflected on the excitability of cortex (Vignaud et al., 2019) so we could detect the difference in EEG between people diagnosed with depression and healthy controls (Ahmadlou et al., 2011; Bachmann et al., 2013; Hosseinifard et al., 2014; Faust et al., 2014; Akar et al., 2015).

It is possible, based on the same nonlinear measures calculated from the resting-state EEG to differentiate between episode and remission of depression (Čukić et al., 2019). To predict clinical outcomes or relapses (for example, after incomplete remission in recurrent depression) would be of great clinical significance especially in clinical psychiatry. A group of authors elucidated risks, pitfalls and recommend the techniques how to improve model reliability and validity in future research (Whelan and Garavan 2013; Gillan and Whelan, 2017; Yahata et al., 2017). All authors described that neuroimaging researchers who start to develop such predictive models are typically unaware of some considerations inevitable to accurately assess model performance and avoid inflated predictions (called 'optimism') (Whelan and Garavan, 2013; Cho et al., 2019). The common characteristics to that kind of research are: classification accuracy is typically overall 80-90%; the size of sample is typically small to modest (less than 50-100 participants); the samples are usually gathered on a single site. Support vector machines (SVM) and its variants are popular but recommendable is the use of embedded regularization frameworks, at least with absolute shrinkage and selection operator (LASO) (Yahata et al., 2017). Leave-one-out and k-fold cross validation are also popular procedures for validation (for model evaluation), and generalization capability of a model is typically untested on an independent sample (Yahata et al., 2017). For model evaluation, Vapnik-Chevronenkis dimension should be used (Vapnik, 1988). A common denominator to majority of studies is a lack of external validation, or even a contamination between training and testing samples. For the sake of methodology we are mentioning here a study which demonstrated an impeccable methodology in machine learning in every aspect, in the task of prediction the responders to medication in MDD (Chekroud et al., 2016). Their algorithms were all written in R.

Generalization is, as we know, the ability of a model that was trained in one dataset to predict patterns in another dataset. When we test the generalizability, we are basically testing whether or not a classification is effective in an independent population. When developing the model, one should be aware of nuisance variables. For example, if using nonlinear measures, they can differ because some of the measures change with age (Goldberger et al., 2000) or they can be characteristic for gender (Ahmadi et al., 2013). It turns out that the algorithm actually learns to recognize that particular dataset with all its characteristics.Overfitting happens when 'a developed model perfectly describes the entire aspects of the training data (including all underlying relationships and associated noise), resulting in fitting error to asymptotically become zero' (Yahata et al., 2017). For that reason, the model will be unable to predict what we want on unseen data (test data).

We must collect more data or establish collaborative project where data can be gathered in numbers which are not achieavable for a single site; a certain standardization of a protocol and decision to share can improve the whole endeavor greatly. Some of great collaborative projects like RDoC, STAR*D, IMAGEN, etc. Also, co-recording with fMRI and MEG should be a solution (Gillan and Dow, 2017). Another line of research is developing wireless EEG caps (Epoch, ENOBIO Neuroelectrics, iMotions, just to mention some) which can be used for research in the environment without restraining the patient, and even for monitoring of recovering from severe episodes. Large-scale imaging campaigns and collection of general population data are the conditions for translation of those research findings to clinics. By regularly allowing their medical data to be a part of such organized collaborative efforts patients would also contribute to the improvement of this precise diagnostic of a (near) future. According to Wessel Kraiij (Data of value, 2017) P4 concept for healthcare improvement stands for: Prediction, Prevention, Personalization and Participation. An important motivation is the observation that healthcare is too focused on disease treatment and not enough on prevention. And another important observation (Kraiij, 2017) is, that treatment and diagnosis are based on population averages. The first project to implement 4P is SWELL project part of a Dutch national ICT program COMMIT (between 2011 and 2016 in the Netherlands, Leiden University). Whelan and Garavan stated that (an unwarranted) 'optimism increases as a function of the decreasing number of participants and the increasing number of predictor variables in the model (model appears better as sample size decreases)' (Whelan and Garavan, 2013). In the same publication they wrote on the importance of keeping training and test subsets completely separate; 'any cross-contamination will result in optimism' (Whelan and Garavan, 2013).

The theory of data mining is clear; all those models work best on bigger samples. Use the repository to test your developed model on the unseen cohort. At least, what we learned is that we need statistics to stop making fools from ourselves (Daniel Takens, 2018). Data mining is the art of finding the meaning from supposedly meaningless data (Peter Flack, 2014). A minimum rate of ten cases per predictor is common (Peduzzi et al., 1996), although not a universal recommendation (Vittinghof and McCulloch, 2007). Optimism can also be lowered with the introduction of the regularization term (Moons et al., 2004). Also, using previous information to constrain model complexity relying on Bayesian approaches is recommendable. Bootstrapping is another helpful method (Efron and Tibshirani, 1993) as well as cross-validation (Efron and Tibshirani, 1997). Cross-validation tests the ability of the model to generalize and involves separating the data into subsets. Both Kohavi and Ng described the technique (Kohavi et al., 1996; Ng et al., 1997). Ng also stated that '…optimism becomes unreliable as the probability of overfitting to the test data increases with multiple comparisons' (Ng et al., 1997).

To conclude we can suggest several levels of standardization in this noble endeavor.

1.     Firstly, the experts in machine learning should establish and maintain high standards in publishing. Practices like registration for the study (or at least registration of the Method) prior the research are not mandatory at the moment, and certain basic requirements must be exercised in order to give us reliable results. Stages like internal and external validation, as well as precautions of contamination of the data would be a must. If journals can exercise those minimal requirements before the submission of manuscripts, the situation would change for the better.

2.     When we get the new piece of equipment in the laboratory it is natural that members of the team first must learn how to use it, how to calibrate it and how to recognize mistakes which can happen during the process. They must learn how that equipment works, what are the limitations of the method used and in general what it is capable of. The same applies to machine learning practices; we must help our students learn how to use it, but for that they have to understand it first and rely on basic postulates of it.

3.     Different disciplines need to develop different standards how to apply certain methods and how to interpret the results of every machine learning method in that particular field. How we test the accuracy will vary from one to another discipline, but the methodology must be defined in order to avoid unwarranted optimism. This is the only way how researchers, professors, reviewers and editors can pursue the decent level of reproducibility in the field.

4.     Lastly, education in the field of machine learning need implementation of some of broader aspects. There are lots of things to be done. We first learn algorithms and how to use tools, but students need to learn more about the particular practical applications end examine them and understand them in appropriate way.